\documentclass[11pt]{article}
\usepackage{times,wrapfig,amsmath,amsfonts,bm,color,enumitem,algorithm,algpseudocode}

\newcommand\tTab{\rule{0pt}{2.5ex}}       
\newcommand\bTab{\rule[-1.5ex]{0pt}{0pt}} 

\newcommand{\inner}[2]{\left\langle #1, #2 \right\rangle}

\newcommand{\E}{\ensuremath{\mathbb{E}}}

\newcommand{\norm}[1]{\left\lVert{#1}\right\rVert}
\newcommand{\abs}[1]{\left\lvert{#1}\right\rvert}

\newcommand{\diag}{\operatorname{diag}}

\newcommand{\R}{\mathbb{R}}

\mathchardef\hyphen="2D

\usepackage{times}
\usepackage{amsmath,amssymb,amsthm}
\usepackage{graphicx}
\usepackage{subfigure}
\usepackage{hyperref}
\usepackage{url}
\usepackage{pbox}

\newcommand{\RR}{\mathbb{R}}

\newtheorem{thm}{Theorem}
\newtheorem{lem}{Lemma}

\newtheorem{Def}{Definition}

\newcommand{\calP}{\mathcal{P}}

\newcommand{\calN}{\mathcal{N}}

\newcommand{\net}{{\rm net}}
\newcommand{\removed}[1]{}

\newcommand{\calT}{\mathcal{T}}
\newcommand{\calA}{\mathcal{A}}
\newcommand{\In}{\text{in}}
\newcommand{\Out}{\text{out}}
\newcommand{\vecX}{\mathbf{x}}

\renewcommand{\vec}[1]{\mathbf{#1}}

\usepackage{wrapfig}
\usepackage{authblk}
\usepackage[utf8]{inputenc} 
\usepackage[T1]{fontenc}    
\usepackage{hyperref}       
\usepackage{url}            
\usepackage{booktabs}       
\usepackage{amsfonts}       
\usepackage{nicefrac}       
\usepackage{microtype}      
\usepackage{caption} 
 \usepackage[margin=1in]{geometry}
\usepackage[outercaption]{sidecap} 

\title{Path-Normalized Optimization of Recurrent Neural Networks\\ with ReLU Activations}

\author[1]{Behnam Neyshabur \footnote{Contributed equally.}}
\author[2]{Yuhuai Wu $^*$}
\author[3]{Ruslan Salakhutdinov}
\author[1]{Nathan Srebro}
\affil[1]{Toyota Technological Institute at Chicago}
\affil[2]{Department of Computer Science, University of Toronto}
\affil[3]{School of Computer Science,  Carnegie Mellon University}

\date{}

\newcommand{\Win}{\vec W_{\text{in}}}
\newcommand{\Wout}{\vec W_{\text{out}}}
\newcommand{\Wrec}{\vec W_{\text{rec}}}
\newcommand{\length}{\text{len}}
\newcommand{\qedwhite}{\hfill \ensuremath{\Box}}

\begin{document}
\maketitle

\begin{abstract}
We investigate the parameter-space geometry of recurrent neural networks (RNNs), 
 and develop an adaptation of path-SGD optimization method, attuned to this geometry, 
that can learn plain RNNs with ReLU activations. 
 On several datasets that require capturing
 long-term dependency structure, we show that path-SGD 
can significantly improve trainability of 
  ReLU RNNs compared to RNNs trained with SGD, 
 even with various recently suggested initialization schemes.
\end{abstract}

\section{Introduction}

Recurrent Neural Networks (RNNs) have been found to be successful in a
variety of sequence learning problems \cite{graves14,cho14,kiros14}, including those involving long
term dependencies (e.g., \cite{arjovsky15,zhang2016architectural}).  However, most of the empirical
success has not been with ``plain'' RNNs but rather with alternate,
more complex structures, such as Long Short-Term Memory (LSTM)
networks \cite{hochreiter97} or Gated Recurrent Units (GRUs) \cite{cho14}.  Much of the
motivation for these more complex models is not so much because
of their modeling richness, but perhaps more
because they seem to be easier to optimize.  As we discuss in Section
\ref{sec:activation}, training plain RNNs using gradient-descent variants seems
problematic, and the choice of the activation function could cause a problem of vanishing gradients or of exploding
gradients.

In this paper our goal is to better understand the geometry of plain RNNs, 
and develop better optimization methods,
adapted to this geometry, that directly learn plain RNNs with ReLU
activations.  One motivation for insisting on plain RNNs, as opposed to
LSTMs or GRUs, is because they are simpler and might be more appropriate for
applications that require low-complexity design such as in mobile
computing platforms \cite{talathi16,han15}.  In other
applications, it might be better to solve
optimization issues by better optimization methods rather than
reverting to more complex models.  Better understanding optimization
of plain RNNs can also assist us in designing, optimizing and
intelligently using more complex RNN extensions.

Improving training RNNs with ReLU activations has been the subject of
some recent attention, with most research focusing on different
initialization strategies \cite{le15,talathi16}.  While initialization can certainly
have a strong effect on the success of the method, it generally can at
most delay the problem of gradient explosion 
during optimization. In this paper we take a different approach
that can be combined with any initialization choice, and focus on the
dynamics of the optimization itself.

Any local search method is inherently tied to some notion of geometry
over the search space (e.g. the space of RNNs). For example, gradient
descent (including stochastic gradient descent) is tied to the
Euclidean geometry and can be viewed as steepest descent with respect
to the Euclidean norm.  Changing the norm (even to a different
quadratic norm, e.g.~by representing the weights with respect to a
different basis in parameter space) results in different optimization
dynamics.  We build on prior work on the geometry and optimization in
feed-forward networks, which uses the path-norm~\cite{NeySalSre15} (defined in Section
\ref{sec:pathsgd}) to determine a geometry leading to the path-SGD optimization
method.  To do so, we investigate the geometry of RNNs as feedforward
networks with shared weights (Section \ref{sec:notation}) and extend a line of
work on Path-Normalized optimization to include networks with
shared weights.  We show that the resulting algorithm (Section \ref{sec:pathsgd})
has similar invariance properties on RNNs as those of standard
path-SGD on feedforward networks, and can result in better
optimization with less sensitivity to the scale of the weights.  

\section{Recurrent Neural Nets as Feedforward Nets with Shared Weights} \label{sec:notation}

We view Recurrent Neural Networks (RNNs) as feedforward networks with shared weights.

We denote a general feedforward network with ReLU activations and shared weights
is indicated by $\calN(G,\pi,\vec p)$ where $G(V,E)$ is a directed
acyclic graph over the set of nodes $V$ that corresponds to units
$v\in V$ in the network, including special subsets of input and output
nodes $V_{\rm in},V_{\rm out}\subset V$, $\vec p\in \RR^m$ is a
parameter vector and $\pi:E\rightarrow \{1,\dots,m\}$ is a mapping
from edges in $G$ to parameters indices.  For any edge $e\in E$, 
the weight of the edge $e$ is indicated by $w_e=p_{\pi(e)}$. We refer to the
set of edges that share the $i$th parameter $p_i$ by $E_i=\left\{ e\in
  E \middle| \pi(e)=i \right\}$.  That is, for any $e_1,e_2\in E_i$,
$\pi(e_1)=\pi(e_2)$ and hence $w_{e_1}=w_{e_2}=p_{\pi(e_1)}$.

Such a feedforward network represents a function $f_{\calN(G,\pi,\vec
  p)}:\RR^{\abs{V_{\rm in}}}\rightarrow\RR^{\abs{V_{\rm out}}}$ as
follows: For any input node $v\in V_{\rm in}$, its output $h_v$ is the
corresponding coordinate of the input vector $\vecX \in
\RR^{\abs{V_{\rm in}}}$. For each internal node $v$, the output is
defined recursively as $h_v=\left[\sum_{(u\rightarrow v)\in E}
  w_{u\rightarrow v}\cdot h_u\right]_+$ where $[z]_+=\max(z,0)$ is the
ReLU activation function\footnote{The bias terms can be modeled by
  having an additional special node $v_{\rm bias}$ that is connected
  to all internal and output nodes, where $h_{v_{\rm bias}}=1$.}. For
output nodes $v\in V_{\rm out}$, no non-linearity is applied and their
output $h_v=\sum_{(u\rightarrow v)\in E} w_{u\rightarrow v}\cdot h_u$
determines the corresponding coordinate of the computed function
$f_{\calN(G,\pi,\vec p)}(\vecX)$.  Since we will fix the graph $G$ and
the mapping $\pi$ and learn the parameters ${\vec p}$, we use the
shorthand $f_{\vec p}=f_{\calN(G,\pi,\vec p)}$ to refer to the function
implemented by parameters ${\vec p}$.  The goal of training is to find
parameters ${\vec p}$ that minimize some error functional $L(f_{\vec p})$ that
depends on ${\vec p}$ only through the function $f_{\vec p}$.  E.g.~in supervised
learning $L(f)=\E\left[\textit{loss}(f(x),y)\right]$ and this is typically done
by minimizing an empirical estimate of this expectation.
        
If the mapping $\pi$ is a one-to-one mapping, then there is no weight
sharing and it corresponds to standard feedforward networks. On the
other hand, weight sharing exists if $\pi$ is a many-to-one mapping.
Two well-known examples of feedforward networks with shared weights
are \emph{convolutional} and \emph{recurrent} networks. We mostly use
the general notation of feedforward networks with shared weights
throughout the paper as this will be more general and simplifies
the development and notation.  However, when focusing on RNNs, it is helpful to
discuss them using a more familiar notation which we briefly introduce
next.

\paragraph{Recurrent Neural Networks}
  
Time-unfolded RNNs are feedforward networks with shared weights that map an input sequence to an output sequence. Each input node corresponds to either a coordinate of the input vector at a particular time step or a hidden unit at time $0$. Each output node also corresponds to a coordinate of the output at a specific time step. Finally, each internal node refers to some hidden unit at time $t\geq 1$. When discussing RNNs, it is useful to refer to different layers and the values calculated at different time-steps. We use a notation for RNN structures in which the nodes are partitioned into layers and $\vec h_t^i$ denotes the output of nodes in layer $i$ at time step $t$. Let $\vecX=(\vecX_1,\dots,\vecX_T)$ be the input at different time steps where $T$ is the maximum number of propagations through time and we refer to it as the length of the RNN. For $0\leq i <d$, let $\Win^i$ and $\Wrec^i$ be the input and recurrent parameter matrices of layer $i$ and $\Wout$ be the output parameter matrix. Table \ref{tab:notation} shows forward computations for RNNs.The output of the function implemented by RNN can then be calculated as $f_{\vec W,t}(x)=h_t^d$. Note that in this notations, weight matrices $\Win$, $\Wrec$ and $\Wout$ correspond to ``free'' parameters of the model that are shared in different time steps.

\begin{table}
\begin{center}
\small
\begin{tabular}{|c|c|c|c|}
\hline
 & Input nodes &  Internal nodes & Output nodes \rule{0pt}{2ex}\\
\hline
FF (shared weights)& $h_v=x[v]$ &  $h_v=\left[\sum_{(u\rightarrow v)\in E} w_{u\rightarrow v} h_u\right]_+$ & $h_v=\sum_{(u\rightarrow v)\in E} w_{u\rightarrow v} h_u$ \bTab \tTab \\
\hline
RNN notation& $\vec h_t^0=\vec x_t,\vec h_0^i=0$& $\vec h_t^i=\left[\Win^i \vec \vec h_t^{i-1} + \Wrec^i \vec \vec h_{t-1}^i\right]_+$& $\vec h_t^d=\Wout \vec h_t^{d-1}$ \bTab \tTab\\
\hline
\end{tabular}
  \caption{\small Forward computations for feedforward nets with shared weights.}
  \label{tab:notation}
 \end{center}
 \vspace{-0.3in}
\end{table}

\section{Non-Saturating Activation Functions} \label{sec:activation}

The choice of activation function for neural networks
can have a large impact on optimization.  We are particularly
concerned with the distinction between ``saturating'' and
``non-starting'' activation functions.  We consider only monotone
activation functions and say that a function is ``saturating'' if it
is bounded---this includes, e.g. sigmoid, hyperbolic tangent and the
piecewise-linear ramp activation functions.  Boundedness necessarily
implies that the function values converge to finite values at negative
and positive infinity, and hence asymptote to horizontal lines on both
sides.  That is, the derivative of the activation converges to zero as
the input goes to both $-\infty$ and $+\infty$.  Networks with
saturating activations therefore have a major shortcoming: the
vanishing gradient problem~\cite{hochreiter98}. The problem here is
that the gradient disappears when the magnitude of the input to an
activation is large (whether the unit is very ``active'' or very
``inactive'') which makes the optimization very challenging.

While sigmoid and hyperbolic tangent have historically been popular
choices for fully connected feedforward and convolutional neural
networks, more recent works have shown undeniable advantages of
non-saturating activations such as ReLU, which is now the standard
choice for fully connected and Convolutional
networks~\cite{nair10,krizhevsky12}. Non-saturating activations,
including the ReLU, are typically still bounded from below and
asymptote to a horizontal line, with a vanishing derivative, at
$-\infty$.  But they are unbounded from above, enabling their
derivative to remain bounded away from zero as the input goes to
$+\infty$.  Using ReLUs enables gradients to not vanish along activated
paths and thus can provide a stronger signal for training.

However, for recurrent neural networks, using ReLU activations is
challenging in a different way, as even a small change in the
direction of the leading eigenvector of the recurrent weights could
get amplified and potentially lead to the explosion in forward or
backward propagation~\cite{arjovsky15}.

To understand this, consider a long path from an input in the first
element of the sequence to an output of the last element, which passes
through the same RNN edge at each step (i.e.~through many edges in some $E_i$ in
the shared-parameter representation).  The length of this path, and
the number of times it passes through edges associated with a single
parameter, is proportional to the sequence length, which could easily
be a few hundred or more.  The effect of this parameter on the path is
therefore exponentiated by the sequence length, as are gradient updates
for this parameter, which could lead to parameter explosion unless an
extremely small step size is used.  

Understanding the geometry of RNNs with ReLUs
could helps us deal with the above issues more effectively. We next
investigate some properties of geometry of RNNs with ReLU activations.

\subsection*{Invariances in Feedforward Nets with Shared Weights}
Feedforward networks (with or without shared weights) are highly
over-parameterized, i.e. there are many parameter settings
${\vec p}$ that represent the same function $f_{\vec p}$.  Since our true object of
interest is the function $f$, and not the identity ${\vec p}$ of the
parameters, it would be beneficial if optimization would depend only
on $f_{\vec p}$ and not get ``distracted'' by difference in ${\vec p}$ that does not
affect $f_{\vec p}$.  It is therefore helpful to study the transformations on
the parameters that will not change the function presented by the
network and come up with methods that their performance is not
affected by such transformations. 

\begin{Def}
We say a network $\calN$ is \emph{invariant} to a transformation $\calT$ if for any parameter setting $\vec p$, $f_{\vec p} =  f_{\calT(\vec p)}$. Similarly, we say an update rule $\calA$ is \emph{invariant} to $\calT$ if for any $\vec p$, $f_{\calA(\vec p)} =  f_{\calA(\calT(\vec p))}$. 
\end{Def}
Invariances have also been studied as different mappings from the parameter space to the same function space~\cite{ollivier2015riemannian} while we define the transformation as a mapping inside a fixed parameter space. A very important invariance in feedforward networks is \emph{node-wise rescaling}~\cite{neyshabur16}. For any internal node $v$ and any scalar $\alpha>0$, we can multiply all incoming weights into $v$ (i.e. $w_{u\rightarrow v}$ for any $(u\rightarrow v)\in E$) by $\alpha$ and all the outgoing weights (i.e. $w_{v\rightarrow u}$ for any $(v\rightarrow u)\in E$) by $1/\alpha$ without changing the function computed by the network. Not all node-wise rescaling transformations can be applied in feedforward nets with shared weights. This is due to the fact that some weights are forced to be equal and therefore, we are only allowed to change them by the same scaling factor. 

\begin{Def}
Given a network $\calN$, we say an invariant transformation $\widetilde{\calT}$ that is defined over edge weights (rather than parameters) is \emph{feasible} for parameter mapping $\pi$ if the shared weights remain equal after the transformation, i.e. for any $i$ and for any $e,e'\in E_i$, $\widetilde{\calT}(\vec w)_e =\widetilde{\calT}(\vec w)_{e'}$. 
\end{Def}

Therefore, it is helpful to understand what are the \emph{feasible} node-wise rescalings for RNNs. In the following theorem, we characterize all feasible node-wise invariances in RNNs.

\begin{SCfigure}
\vspace{-11pt}
\includegraphics[width=10cm]{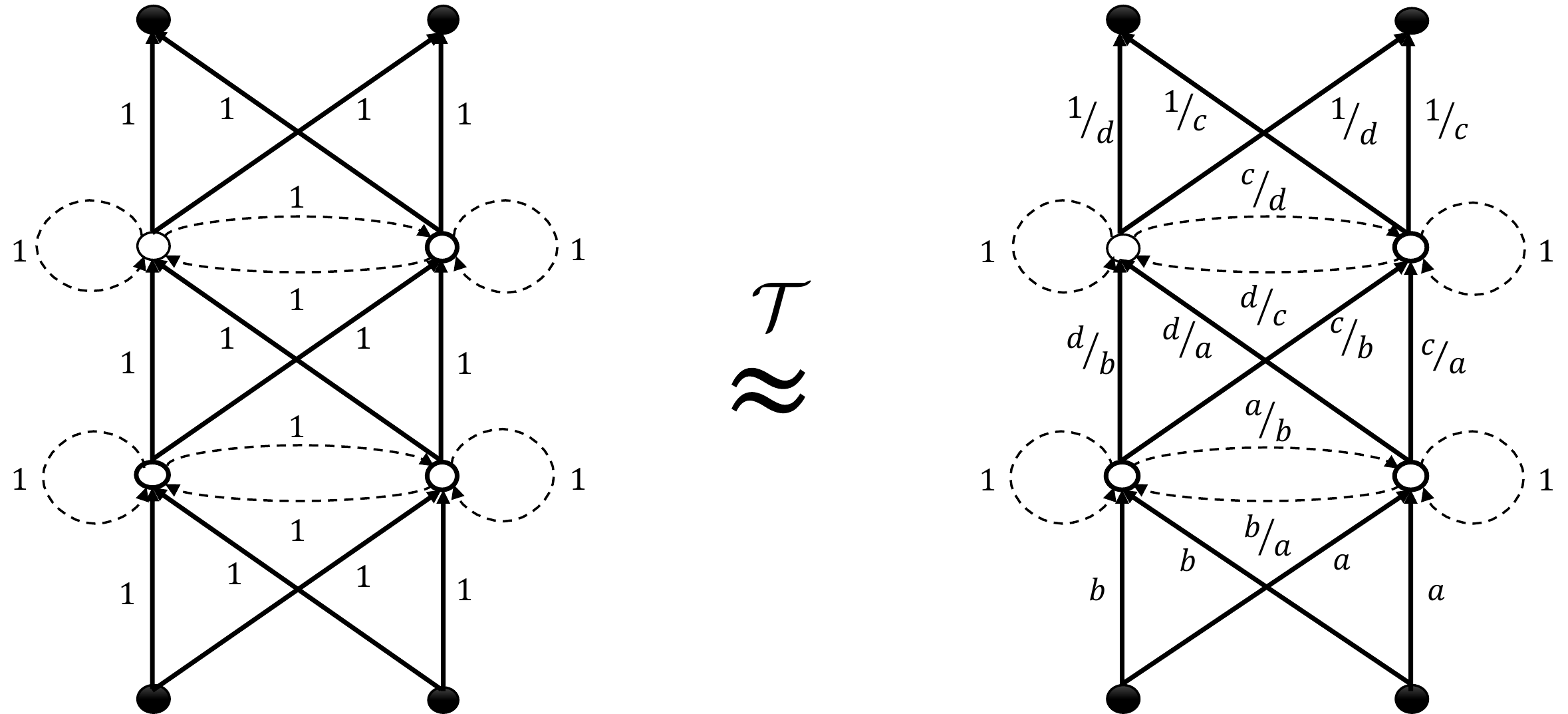}
\caption{\small An example of invariances in an RNN with two hidden layers each of which has 2 hidden units. The dashed lines correspond to recurrent weights. The network on the left hand side is equivalent (i.e. represents the same function) to the network on the right for any nonzero $\alpha^1_1=a$, $\alpha^1_2=b$, $\alpha^2_1=c$, $\alpha^2_2=d$. }
\label{fig:invariance}
\end{SCfigure}

\begin{thm}\label{thm:net-inv}
For any $\alpha$ such that $\alpha^i_j>0$, any Recurrent Neural Network with ReLU activation is invariant to the transformation
$\calT_\alpha\left(\left[\Win,\Wrec,\Wout\right]\right) = \left[\calT_{\In,\alpha}\left(\Win\right),\calT_{\text{rec},\alpha}\left(\Wrec\right),\calT_{\Out,\alpha}\left(\Wout\right)\right]$ where for any $i,j,k$:
\begin{align}
&\calT_{\In,{\alpha}}( \Win)^{i}[j,k]= 
\begin{cases}
\alpha^i_j \Win^{i}[j,k] & i=1, \\ 
\left(\alpha^i_j /\alpha^{i-1}_k \right) \Win^{i}[j,k] & 1< i <d,\\
\end{cases}\\
\calT_{\text{rec},\alpha}( \Wrec )^{i}[j,k]&=
\left(\alpha^i_j /\alpha^{i}_k \right) \Wrec^{i}[j,k],\qquad \calT_{\Out,{\alpha}}( \Wout )[j,k]=\left(1 /\alpha^{d-1}_k \right) \Wout[j,k]. \notag
\end{align}
Furthermore, any feasible node-wise rescaling transformation can be presented in the above form.
\end{thm}

The proofs of all theorems and lemmas are given in Appendix~\ref{sec:proofs}. The above theorem shows that there are many transformations under which RNNs represent the same function. An example of such invariances is shown in Fig.~\ref{fig:invariance}. Therefore, we would like to have optimization algorithms that are invariant to these transformations and in order to do so, we need to look at measures that are invariant to such mappings.

\section{Path-SGD for Networks with Shared Weights} \label{sec:pathsgd}

As we discussed, optimization is inherently tied to a choice of
geometry, here represented by a choice of complexity measure or
``norm''\footnote{The path-norm which we define is a norm on
  functions, not on weights, but as we prefer not getting into this
  technical discussion here, we use the term ``norm'' very loosely to
  indicate some measure of magnitude~\cite{NeyTomSre15}.}.  Furthermore, we prefer using
an invariant measure which could then lead to an invariant
optimization method.  In Section \ref{sec:path-reg} we introduce the
path-regularizer and in Section \ref{sec:path-forward}, the derived Path-SGD
optimization algorithm for standard feed-forward networks.  Then in
Section \ref{sec:path-shared} we extend these notions also to networks with shared
weights, including RNNs, and present two invariant optimization
algorithms based on it.  In Section \ref{sec:path-compute} we show how these can be
implemented efficiently using forward and backward propagations.

\subsection{Path-regularizer}\label{sec:path-reg}
The path-regularizer is the sum over all paths from input nodes to
output nodes of the product of squared weights along the path. To define
it formally, let $\calP$ be the set of directed paths from input to
output units so that for any path
$\zeta=\left(\zeta_0,\dots,\zeta_{\length(\zeta)}\right)\in \calP$ of
length $\length(\zeta)$, we have that $\zeta_0\in V_\In$,
$\zeta_{\length(\zeta)}\in V_\Out$ and for any $0\leq i \leq
\length(\zeta)-1$, $(\zeta_{i}\rightarrow \zeta_{i+1})\in E$. We also
abuse the notation and denote $e\in \zeta$ if for some $i$,
$e=(\zeta_i,\zeta_{i+1})$. Then the path regularizer can be written
as:
\begin{equation}
\gamma_\net^2(\vec w) = \sum_{\zeta \in \calP} \prod_{i=0}^{\length(\zeta)-1} w_{\zeta_{i}\rightarrow \zeta_{i+1}}^2
\end{equation}
Equivalently, the path-regularizer can be defined recursively on the
nodes of the network as:
\begin{equation}
\gamma_v^2(\vec w) = \sum_{(u\rightarrow v)\in E} \gamma^2_u(\vec w) w_{u \rightarrow v}^2\;, \qquad \gamma_\net^2(\vec w) = \sum_{u\in V_\Out} \gamma^2_u(\vec w)
\end{equation}

\subsection{Path-SGD for Feedforward Networks} \label{sec:path-forward}

Path-SGD is an approximate steepest descent step with respect to the
path-norm.  More formally, for a network without shared weights, where
the parameters are the weights themselves, consider the diagonal
quadratic approximation of the path-regularizer about the current
iterate $\vec w^{(t)}$:
\begin{equation}
\hat{\gamma}^2_{\rm net}(\vec w^{(t)}+\Delta \vec w) = \gamma^2_{\rm net}(\vec w^{(t)})+
\inner{\nabla \gamma^2_{\rm net}(\vec w^{(t)})}{\Delta \vec w} +
\frac{1}{2} \Delta \vec w^\top
\diag\left(\nabla^2  \gamma^2_{\rm net}(\vec w^{(t)})\right) \Delta \vec w
\end{equation}
Using the corresponding quadratic norm $\norm{\vec w-\vec w'}_{\hat{\gamma}^2_{\rm net}(\vec w^{(t)}+\Delta \vec w)}^2=\frac{1}{2}\sum_{e\in E}\frac{\partial^2 \gamma^2_{\rm 
net}}{\partial w^2_{e}}\left(w_e-w'_e\right)^2$, we can define an
approximate steepest descent step as:
\begin{equation}\label{eq:stp}
\vec w^{(t+1)}=\min_{\vec w} \eta \inner{\nabla L(\vec w)}{\vec w-\vec w^{(t)}} + \norm{\vec w-\vec w^{(t)}}_{\hat{\gamma}^2_{\rm net}(\vec w^{(t)}+\Delta \vec w)}^2.
\end{equation}
Solving \eqref{eq:stp} yields the
update:
\begin{equation}\label{eq:update-forward}
w^{(t+1)}_{e} = w^{(t)}_{e} - \frac{\eta}{\kappa_{e}(\vec w^{(t)})} \frac{\partial 
L}{\partial w_{e}}(\vec w^{(t)}) \quad\quad\textrm{where: } \kappa_{e}(\vec w)=\frac{1}{2}\frac{\partial^2 \gamma^2_{\rm 
net}(\vec w)}{\partial w^2_{e}}.
\end{equation}
The stochastic version that uses a subset
of training examples to estimate $\frac{\partial
  L}{\partial w_{u\rightarrow v}}(\vec w^{(t)})$ is called Path-SGD
\cite{NeySalSre15}. We now show how Path-SGD can be extended to networks with
shared weights.

\subsection{Extending to Networks with Shared Weights} \label{sec:path-shared}
When the networks has shared weights, the path-regularizer is a
function of parameters $\vec p$ and therefore the quadratic
approximation should also be with respect to the iterate $\vec
p^{(t)}$ instead of $\vec w^{(t)}$ which results in the following
update rule:
\begin{equation}\label{eq:stp-sh}
\vec p^{(t+1)}=\min_{\vec p} \eta \inner{\nabla L(\vec p)}{\vec p-\vec p^{(t)}}+ \norm{\vec p-\vec p^{(t)}}_{\hat{\gamma}^2_{\rm net}(\vec p^{(t)}+\Delta \vec p)}.
\end{equation}
where $\norm{\vec p-\vec p'}_{\hat{\gamma}^2_{\rm net}(\vec p^{(t)}+\Delta \vec p)}^2=\frac{1}{2}\sum_{i=1}^m\frac{\partial^2 \gamma^2_{\rm 
net}}{\partial p^2_{i}}\left(p_i-p'_i\right)^2$. Solving \eqref{eq:stp-sh} gives the following update:
\begin{equation}\label{eq:update-shared}
p^{(t+1)}_{i} = p^{(t)}_{i} - \frac{\eta}{\kappa_{i}(\vec p^{(t)})} \frac{\partial 
L}{\partial p_{i}}(\vec p^{(t)}) \quad\quad\textrm{where: } \kappa_{i}(\vec p)=\frac{1}{2}\frac{\partial^2 \gamma^2_{\rm 
net}(\vec p)}{\partial p^2_{i}}.
\end{equation}
The second derivative terms $\kappa_i$ are specified in terms of their
path structure as follows:
\begin{lem}\label{lem:path} $\kappa_{i}(\vec p) = \kappa^{(1)}_{i}(\vec p) + \kappa^{(2)}_{i}(\vec p)$
where
\begin{align}
\kappa^{(1)}_{i}(\vec p) &= \sum_{e\in E_i}\sum_{\zeta \in
    \calP} \vec 1_{e\in \zeta} \prod_{j=0 \atop e\neq
    (\zeta_j\rightarrow \zeta_{j+1})}^{\length(\zeta)-1}
  p^2_{\pi(\zeta_j\rightarrow \zeta_{j+1})}= \sum_{e\in E_i} \kappa_{e}(\vec w), \label{eq:k1}\\
 \kappa^{(2)}_{i}(\vec p)  &= p_i^2\sum_{e1,e2\in E_i \atop e_1\neq e_2}\sum_{\zeta \in \calP} \vec 1_{e_1,e_2\in \zeta} \prod_{j=0 \atop { e_1\neq (\zeta_j\rightarrow \zeta_{j+1}) \atop e_2\neq (\zeta_j\rightarrow \zeta_{j+1}) }}^{\length(\zeta)-1} p^2_{\pi(\zeta_j\rightarrow \zeta_{j+1})}, 
 \label{eq:k2}
\end{align}
and $\kappa_e(\vec w)$ is defined in \eqref{eq:update-forward}.
\end{lem}
The second term $\kappa^{(2)}_i(\vec p)$ measures the effect of
interactions between edges corresponding to the same parameter (edges
from the same $E_i$) on the same path from input to output.  In
particular, if for any path from an input unit to an output unit, no
two edges along the path share the same parameter, then $
\kappa^{(2)}(\vec p)=0$. For example, for any feedforward or
Convolutional neural network, $\kappa^{(2)}(\vec p)=0$.  But for RNNs,
there certainly are multiple edges sharing a single parameter on the
same path, and so we could have $\kappa^{(2)}(\vec p)\neq 0$.


The above lemma gives us a precise update rule for the approximate
steepest descent with respect to the path-regularizer. The following
theorem confirms that the steepest descent with respect to this
regularizer is also invariant to all feasible node-wise rescaling for
networks with shared weights.

\begin{thm}\label{thm:pathsgd-inv}
  For any feedforward networks with shared weights, the update
  \eqref{eq:update-shared} is invariant to all feasible node-wise rescalings.
  Moreover, a simpler update rule that only uses $\kappa^{(1)}_i(\vec
  p)$ in place of $\kappa_i(\vec p)$ is also invariant to all feasible
  node-wise rescalings.
\end{thm}

Equations \eqref{eq:k1} and \eqref{eq:k2} involve a sum over all paths
in the network which is exponential in depth of the network.
However, we next show that both of these equations can be calculated
efficiently.

\subsection{Simple and Efficient Computations for RNNs} \label{sec:path-compute}
We show how to calculate $\kappa^{(1)}_i(\vec p)$ and
$\kappa^{(2)}_i(\vec p)$ by considering a network with the
same architecture but with squared weights:
\begin{thm}\label{thm:pathsgd-cal}
For any network $\calN(G,\pi,p)$, consider $\calN(G,\pi,\tilde{p})$ where for any $i$, $\tilde{p}_i=p_i^2$. Define the function $g:\R^{\abs{V_\In}}\rightarrow \R$ to be the sum of outputs of this network: $g(x)=\sum_{i=1}^{\abs{V_{\Out}}}f_{\tilde{\vec p}}(x)[i]$. Then $\kappa^{(1)}$ and $\kappa^{(2)}$ can be calculated as follows where $\vec 1$ is the all-ones input vector:
\begin{equation}
\kappa^{(1)}(\vec p) = \nabla_{\tilde{\vec p}} g(\vec 1),\qquad \kappa^{(2)}_i(\vec p)  = \sum_{(u\rightarrow v),(u'\rightarrow v')\in E_i \atop{(u\rightarrow v) \neq (u'\rightarrow v')} } \tilde{p}_i\frac{\partial g(\vec 1)}{\partial h_{v'}(\tilde{\vec p})}\frac{\partial h_{u'}(\tilde{\vec p})}{\partial h_v(\tilde{\vec p})} h_u(\tilde{\vec p}).
\end{equation}
\end{thm}
In the process of calculating the gradient $\nabla_{\tilde{\vec p}}
g(\vec 1)$, we need to calculate $h_u(\tilde{\vec p})$ and $\partial g(\vec
1)/\partial h_{v}(\tilde{\vec p})$ for any $u,v$. Therefore, the only
remaining term to calculate (besides $\nabla_{\tilde{p}} g(\vec 1)$) is
$\partial h_{u'}(\tilde{\vec p})/\partial h_v(\tilde{\vec p})$.

Recall that $T$ is the length (maximum number of propagations through time) and $d$ is the number 
of layers in an RNN. Let $H$ be the number of hidden units in each layer and $B$
be the size of the mini-batch. Then calculating the gradient of the
loss at all points in the minibatch (the standard work required for
any mini-batch gradient approach) requires time $O(BdTH^2)$. In order
to calculate $\kappa^{(1)}_i(\vec p)$, we need to calculate the gradient
$\nabla_{\tilde{\vec p}}g(1)$ of a similar network at a {\em single}
input---so the time complexity is just an additional $O(dTH^2)$. The second term
$\kappa^{(2)}(\vec p)$ can also be calculated for RNNs in $O(dTH^2(T+H))$ \footnote{
For an RNN, $\kappa^{(2)}(\Win)=0$ and $\kappa^{(2)}(\Wout)=0$ because only
recurrent weights are can be shared multiple times along an input-output path. 
$\kappa^{(2)}(\Wrec)$ can be written and calculated in the matrix form:
{\tiny$\kappa^{(2)}(\Wrec^i) =\Wrec'^i \odot  \sum_{t_1=0}^{T-3}\left[\left(\left(\Wrec'^i\right)^{t_1}\right)^\top \odot \sum_{t_2=2}^{T-t_1-1} 
\frac{\partial g(\vec 1)}{\partial \vec h^i_{t_1+t_2+1}(\tilde{\vec p})} \left(\vec h^i_{t_2}(\tilde{\vec p})\right)^\top \right]
$}
where for any $i,j,k$ we have $\Wrec'^i[j,k] = \left(\Wrec^i[j,k]\right)^2$. 
The only terms that require extra computation are powers of
$\Wrec$ which can be done in $O(dTH^3)$ and the rest of the matrix
computations need $O(dT^2H^2)$.}.
Therefore, the ratio of time complexity of calculating the
first term and second term with respect to the gradient over
mini-batch is $O(1/B)$ and $O((T+H)/B)$ respectively.  Calculating
only $\kappa^{(1)}_i(\vec p)$ is therefore very cheap with minimal
per-minibatch cost, while calculating $\kappa^{(2)}_i(\vec p)$ might
be expensive for large networks.  Beyond the low computational cost,
calculating $\kappa^{(1)}_i(\vec p)$ is also very easy to implement as
it requires only taking the gradient with respect to a standard
feed-forward calculation in a network with slightly modified
weights---with most deep learning libraries it can be implemented very
easily with only a few lines of code.


\section{Experiments} \label{sec:experiments}

\removed{
\begin{table}
\begin{center}
\small
\begin{tabular}{|c|c|c|c|}
\hline
 &  $T=10$ & $T=20$ & $T=40$\\
\hline
$H=100$ &  0.00037&  0.00044& 0.00048\\
\hline
$H=200$ &  0.00027&  0.00029& 0.00032\\
\hline
$H=400$  &  0.00014&  0.00016& 0.00022\\
\hline
\end{tabular}
  \caption{\small The ratio $\norm{\kappa^{(2)}}_2/\norm{\kappa^{(1)}}_2$ for different lengths $T$ and number of hidden units $H$. }
  \label{tab:compare}
 \end{center}
\end{table}

}

\subsection{The Contribution of the Second Term}
\begin{figure*}[t!]
        \centering
        \includegraphics[height=1.5in]{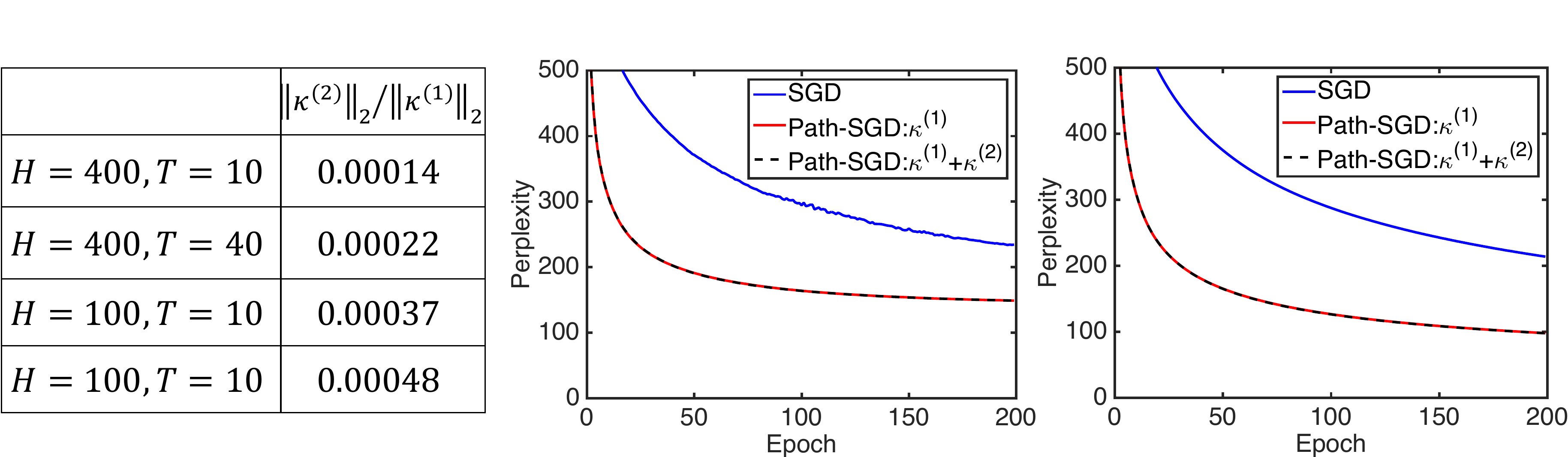}
    \caption{\small Path-SGD with/without the second term in word-level language modeling on PTB. We use the standard split (929k training, 73k validation and 82k test) and the vocabulary size of 10k words. We initialize the weights by sampling from the uniform distribution with range $[-0.1,0.1]$. The table on the left shows the ratio of magnitude of first and second term for different lengths $T$ and number of hidden units $H$. The plots compare the training and test errors using a mini-batch of size 32 and backpropagating through $T=20$ time steps and using a mini-batch of size 32 where the step-size is chosen by a grid search.}
    \label{fig:path-vs-sgd}
    
\begin{picture}(0,0)(0,0)
{\put(120,190){\tiny Test Error}\put(-12,190){\tiny Training Error}}
\end{picture}
\vspace{-0.2in}
\end{figure*}
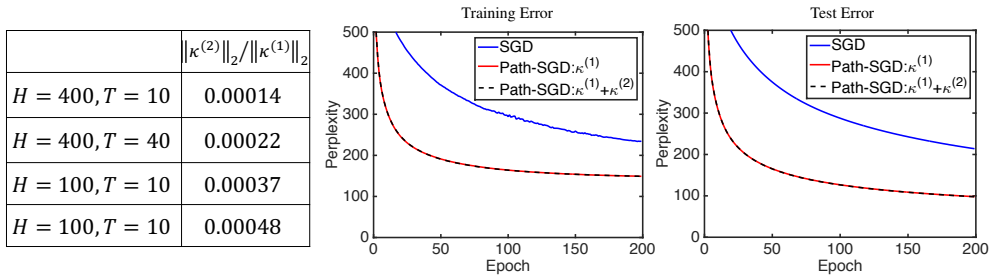

As we discussed in section \ref{sec:path-compute}, the second term $\kappa^{(2)}$ in the update rule can be computationally expensive for large networks. In this section
we investigate the significance of the second term and show that at least in our experiments, the contribution of the second term is negligible.
To compare the two terms $\kappa^{(1)}$ and $\kappa^{(2)}$, we train a single layer RNN with $H=200$ hidden units for the task of word-level language modeling on Penn Treebank (PTB) Corpus~\cite{marcus1993}. Fig.~\ref{fig:path-vs-sgd} compares the performance of SGD vs. Path-SGD with/without  $\kappa^{(2)}$. We clearly see that both version of Path-SGD are performing very similarly and both of them outperform SGD significantly. This results in Fig.~\ref{fig:path-vs-sgd} suggest that the first term is more significant and therefore we can ignore the second term.

To better understand the importance of the two terms, we compared 
the ratio of the norms $\norm{\kappa^{(2)}}_2/\norm{\kappa^{(1)}}_2$ for different RNN lengths $T$ and number of hidden units $H$. 
The table in Fig.~\ref{fig:path-vs-sgd} shows that the contribution of the second term is bigger when the network has fewer number of hidden units and the length of the RNN is larger ($H$ is small and $T$ is large). However, in many cases, it appears that the first term has a much bigger contribution in the update step and hence the second term can be safely ignored. Therefore, in the rest of our experiments, we calculate the Path-SGD updates only using the first term $\kappa^{(1)}$.

\subsection{Synthetic Problems with Long-term Dependencies}

Training Recurrent Neural Networks is known to be hard for modeling long-term dependencies due to the 
gradient vanishing/exploding problem \cite{hochreiter98,bengio1994learning}. In this section, we consider synthetic problems that are specifically designed to test the ability of a model to capture the long-term dependency structure. Specifically, we consider the addition problem and the sequential MNIST problem. 

{\bf Addition problem}: 
The addition problem was introduced in \cite{hochreiter97}. Here,
each input consists of two sequences of length $T$, one of which includes numbers sampled from the uniform distribution with range $[0, 1]$ and the other sequence serves as a mask which is filled with zeros except for two entries.
These two entries indicate which of the two numbers in the first sequence we need to add and the task is to output the result of this addition.\\ 
\textbf{Sequential MNIST}:
In sequential MNIST, each digit image is reshaped into a sequence of length $784$,
turning the digit classification task into sequence classification 
with long-term dependencies~\cite{le15,arjovsky15}.

For both tasks, we closely follow the experimental protocol in \cite{le15}. 
We train a single-layer RNN consisting of 100 hidden units with path-SGD, referred to as \textbf{RNN-Path}. We also train an RNN of the same size with identity initialization, as was proposed in~\cite{le15}, using SGD as our baseline model, referred to as \textbf{IRNN}. We performed grid search for the
learning rates over $\{10^{-2},10^{-3},10^{-4}\}$ for both our model and the baseline.  Non-recurrent weights were initialized from the uniform distribution with range $[-0.01,0.01]$.
Similar to~\cite{arjovsky15}, we found the IRNN to be fairly unstable (with SGD optimization typically diverging). Therefore for IRNN, we ran 10 different initializations and picked the one that did not explode to show its performance. 

\begin{figure}[t]
\centering
\includegraphics[width=15cm]{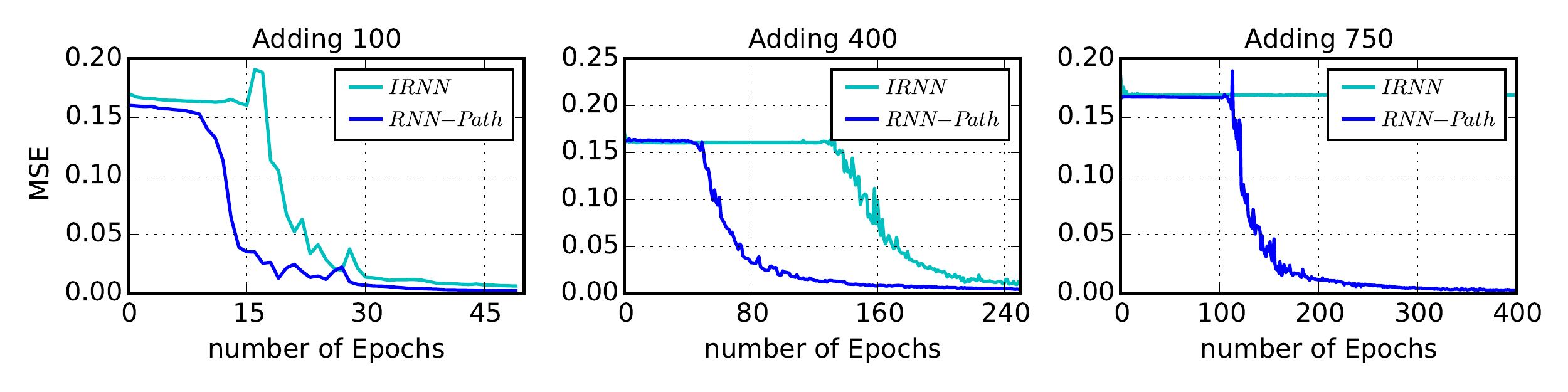}
\vspace{-0.15in}
\caption{\small Test errors for the addition problem of different lengths.}
\label{fig:adding}
\end{figure}

In our first experiment, we evaluate Path-SGD on the addition problem. 
The results are shown in Fig.~\ref{fig:adding} with increasing the length $T$ of the sequence: 
$\{100,400,750\}$. We note that this problem becomes much harder as $T$ increases because the dependency between
the output (the sum of two numbers) and the corresponding inputs becomes more distant. 
We also compare RNN-Path with the previously published results, 
including identity initialized RNN ~\cite{le15} (IRNN), unitary RNN \cite{arjovsky15} (uRNN), and np-RNN\footnote{The original paper does not include any result for 750, so we implemented np-RNN for comparison. However, in our implementation the np-RNN is not able to 
even learn sequences of length of 200. Thus we put ``>2'' for length of 750.} introduced by \cite{talathi16}. Table \ref{tb:adding_mnist} shows the effectiveness of using Path-SGD. Perhaps 
more surprisingly,
with the help of path-normalization, a simple RNN with the identity initialization is able to achieve a 0\% error on the sequences of length 750, 
whereas all the other methods, including LSTMs, fail. This shows that Path-SGD may help stabilize the training and alleviate the gradient problem, so as to perform well on longer sequence. 
We next tried to model the sequences length of 1000, but we found that for such very long 
sequences RNNs, even with Path-SGD, fail to learn. 

\begin{table}[t!]
    \begin{minipage}{.5\linewidth}
      \centering
      \footnotesize
        \begin{tabular}{c c c c c}
    \toprule
    &Adding & Adding& Adding  & \\
        & 100 & 400& 750  &   sMNIST\\
    \midrule
    IRNN~\cite{le15}& 0& 16.7& 16.7 & 5.0\\
    uRNN \cite{arjovsky15} & 0 & 3 & 16.7 & 4.9 \\
    LSTM \cite{arjovsky15}  & 0 & 2 & 16.7 & 1.8\\
    np-RNN\cite{talathi16}        & 0 & 2 & >2 & 3.1\\
    \midrule
    IRNN & 0 & 0 & 16.7  &7.1\\
    RNN-Path & 0& 0 & 0 &  3.1\\
    \bottomrule
  \end{tabular}
        \caption{\small Test error (MSE) for the adding problem with different input sequence lengths and test classification error for the sequential MNIST.}
        \label{tb:adding_mnist}
      
    \end{minipage}%
    \;\;
    \begin{minipage}{.5\linewidth}
    \footnotesize
      \centering
       \begin{tabular}{lcc}
    \toprule
        & PTB & text8 \\
    \midrule
    RNN+smoothReLU \cite{pachitariu2013regularization} & - & 1.55\\
    HF-MRNN \cite{mikolov2012subword}    &  1.42 & 1.54\\
    RNN-ReLU\cite{KruegerM15} & 1.65 & -\\
    RNN-tanh\cite{KruegerM15} & 1.55 & -\\
    TRec,$\beta=500$\cite{KruegerM15} & 1.48 & -\\
    \midrule
    RNN-ReLU & 1.55 & 1.65\\
    RNN-tanh & 1.58 & 1.70\\
    RNN-Path & 1.47 & 1.58\\
    LSTM  & 1.41 & 1.52\\
    \bottomrule
  \end{tabular}
        \caption{\small Test BPC for PTB and text8.}
        \label{tb:bpc_penn_text8}
    \end{minipage} 
\end{table}

Next, we evaluate Path-SGD on the Sequential MNIST problem. 
Table \ref{tb:adding_mnist}, right column, reports test error rates achieved by RNN-Path 
compared to the previously published results. Clearly,
using Path-SGD helps RNNs achieve better generalization. In many cases,
RNN-Path outperforms other RNN methods (except for LSTMs), even for such a long-term dependency problem.

\subsection{Language Modeling Tasks}

In this section we evaluate Path-SGD on a language modeling task. 
We consider two datasets, Penn Treebank (PTB-c) and text8~\footnote{http://mattmahoney.net/dc/textdata}.
\textbf{PTB-c}: We performed experiments on a tokenized Penn Treebank Corpus, following 
the experimental protocol of~\cite{KruegerM15}. The training, validations and test data contain 5017k, 393k and 442k characters respectively.
The alphabet size is 50, and each training sequence is of length 50. \textbf{text8}: The text8 dataset contains 100M characters from Wikipedia with an alphabet size of 27. We follow the data partition of~\cite{mikolov2012subword}, 
where each training sequence has a length of 180.
Performance is evaluated using bits-per-character (BPC) metric, which is $\log_2$ of perplexity. 

Similar to the experiments on the synthetic datasets, 
for both tasks, we train a single-layer RNN consisting of 2048 hidden units with path-SGD (RNN-Path).
Due to the large dimension of hidden space, SGD can take a fairly long time to converge. 
Instead, we use Adam optimizer~\cite{kingma2014adam} to help speed up the training, where 
we simply use the path-SGD gradient as input to the Adam optimizer. 

We also train three additional baseline models: a ReLU RNN with 2048 hidden units, a tanh RNN with 2048 hidden units, 
and an LSTM with 1024 hidden units, 
all trained using Adam. We performed grid search for learning rate over $\{10^{-3},5\cdot10^{-4},10^{-4}\}$ for all of our models.  For ReLU RNNs, we initialize the recurrent matrices 
from uniform$[-0.01,0.01]$, and uniform$[-0.2,0.2]$ for non-recurrent weights. 
For LSTMs, we use orthogonal initialization~\cite{saxe2013exact} for the recurrent matrices and 
uniform$[-0.01,0.01]$ for non-recurrent weights. The results are summarized in Table \ref{tb:bpc_penn_text8}.

\removed{
\begin{table}[t]
  \footnotesize
  \centering
  \begin{tabular}{ll}
    \toprule
    Penn-Treebank     & BPC     \\
    \midrule
    RNN+smoothReLU \cite{pachitariu2013regularization} & - & 1.55\\
    HF-MRNN \cite{mikolov2012subword}    &  1.42 & 1.54\\
    RNN-ReLU\cite{KruegerM15} & 1.65 & -\\
    RNN-tanh\cite{KruegerM15} & 1.55 & -\\
    TRec,$\beta=500$\cite{KruegerM15} & 1.48 & -\\
    \midrule
    RNN-ReLU & 1.55 & 1.65\\
    RNN-tanh & 1.58 & 1.70\\
    RNN-Path & 1.47 & 1.58\\
    LSTM  & 1.41 & 1.52\\
    \bottomrule
  \end{tabular}
  \quad
  \begin{tabular}{ll}
    \toprule
    text8     & BPC     \\
    \midrule
    RNN+smoothReLU \cite{pachitariu2013regularization}  &1.55  \\
    HF-MRNN \cite{mikolov2012subword}    &  1.54\\
    \midrule
    RNN-ReLU & 1.65 \\
    RNN-tanh & 1.70 \\
    RNN-Path & 1.58 \\
    LSTM  & 1.52\\
    \bottomrule
  \end{tabular}
\caption{\small Left: test Bits Per characters on PTB.
Right: test Bits Per characters on text8.
}
\label{tb:bpc_penn_text8}
\end{table}

}

We also compare our results to an RNN that uses 
hidden activation regularizer~\cite{KruegerM15} (TRec,$\beta=500$), 
Multiplicative RNNs trained by Hessian Free methods~\cite{mikolov2012subword}
(HF-MRNN), and an RNN with smooth version of ReLU \cite{pachitariu2013regularization}.
Table~\ref{tb:bpc_penn_text8} shows that path-normalization is able to 
outperform RNN-ReLU and RNN-tanh,
while at the same time 
shortening the performance gap between plain RNN and other more complicated models (e.g. LSTM by 57\% on PTB and 54\% on text8 datasets).
This demonstrates the efficacy of path-normalized optimization for training RNNs with ReLU activation.

\section{Conclusion}
We investigated the geometry of RNNs in a broader class of feedforward networks with shared weights and  showed how understanding the geometry can lead to significant improvements on different learning tasks. Designing an optimization algorithm with a geometry that is well-suited for RNNs, we closed over half of the performance gap between vanilla RNNs and LSTMs. This is particularly useful for applications in which we seek compressed models with fast prediction time that requires minimum storage; and also a step toward bridging the gap between LSTMs and RNNs.

\subsubsection*{Acknowledgments}
This research was supported in part by an NSF RI-AF award and by Intel ICRI-CI. We thank Saizheng Zhang for sharing a base code for RNNs.

\bibliographystyle{plain}
\bibliography{ref}

\newpage
\appendix


\section{Proofs}\label{sec:proofs}
\subsection{Proof of Theorem~\ref{thm:net-inv}}
We first show that any RNN is invariant to $\calT_\alpha$ by induction on layers and time-steps. More specifically, we prove that for any $0\leq t \leq T$ and $1\leq i<d$, $\vec h_t^i\left(\calT_\alpha(\vec W)\right)[j] = \alpha^i_j\vec h_t^i(\vec W)[j]$. The statement is clearly true for $t=0$; because for any $i,j$, $\vec h_0^i\left(\calT_\alpha(\vec W)\right)[j] = \alpha^i_j\vec h_0^i(\vec W)[j]=0$.

Next, we show that for $i=1$, if we assume that the statement is true for $t=t'$, then it is also true for $t=t'+1$:
\begin{align*}
\vec h_{t'+1}^1\left(\calT_\alpha(\vec W)\right)[j] &= \left[\sum_{j'}\calT_{\In,\alpha}(\Win)^1[j,j']\vec x_{t'+1}[j']+ \calT_{\text{rec},\alpha}(\Wrec)^1[j,j'] \vec h_{t'}^1\left(\calT_\alpha(\vec W)\right)[j']\right]_+\\
&=\left[\sum_{j'}\alpha^1_j\Win^1[j,j']\vec x_{t'+1}[j'] +\left(\alpha^1_j /\alpha^1_{j'} \right) \Wrec^1[j,j'] \alpha^1_{j'}\vec h_{t'}^1(\vec W))[j']\right]_+\\
&= \alpha^1_j\vec h_t^i(\vec W)[j]
\end{align*}

We now need to prove the statement for $1<i<d$. Assuming that the statement is true for $t\leq t'$ and the layers before $i$, we have:
\begin{align*}
\vec h_{t'+1}^i\left(\calT_\alpha(\vec W)\right)[j] &= \left[\sum_{j'}\calT_{\In,\alpha}(\Win)^i[j,j']\vec h_{t'+1}^{i-1}\left(\calT_\alpha(\vec W)\right)[j'] +\calT_{\text{rec},\alpha}(\Wrec)^i[j,j'] \vec h_{t'}^i\left(\calT_\alpha(\vec W)\right)[j']\right]_+\\
&=\left[\sum_{j'}\frac{\alpha^i_j}{\alpha^{i-1}_{j'}}\Win^i[j,j']\alpha^{i-1}_{j'}\vec h_{t'+1}^{i-1}(\vec W))[j'] + \frac{\alpha^i_j }{ \alpha^i_{j'} }\Wrec^i[j,j'] \alpha^i_{j'}\vec h_{t'}^i(\vec W))[j']\right]_+\\
&= \alpha^i_j\vec h_t^i(\vec W)[j]
\end{align*}
Finally, we can show that the output is invariant for any $j$ at any time step $t$:
\begin{align*}
f_{\calT(\vec W),t}(\vec x_t)[j] &=  \sum_{j'} \calT_{\text{out},\alpha}(\Wout)[j,j'] \vec h_{t}^{d-1}(\calT_{\alpha}(\vec W)[j']=\sum_{j'} (1/\alpha^{d-1}_{j'})\Wout[j,j'] \alpha^{d-1}_{j'}\vec h_{t}^{d-1}(\vec W)[j'] \\
&=\sum_{j'}\Wout[j,j'] \vec h_{t}^{d-1}(\vec W)[j'] = f_{\vec W,t}(\vec x_t)[j]
\end{align*}

We now show that any feasible node-wise rescaling can be presented as $\calT_\alpha$. Recall that node-wise rescaling invariances for a general feedforward network can be written as $\widetilde{\calT_\beta}(\vec w)_{u\rightarrow v} = (\beta_v/\beta_u)w_{u\rightarrow v}$ for some $\beta$ where $\beta_v>0$ for internal nodes and $\beta_v=1$ for any input/output nodes. An RNN with $T=0$ has no weight sharing and for each node $v$ with index $j$ in layer $i$, we have $\beta_v=\alpha_j^i$. For any $T>0$ however, we there is no invariance that is not already counted. The reason is that by fixing the values of $\beta_v$ for the nodes in time step 0, due to the feasibility, the values of $\beta$ for nodes in other time-steps should be tied to the corresponding value in time step $0$. Therefore, all invariances are included and can be presented in form of $\calT_\alpha$.

 \qedwhite

\subsection{Proof of Lemma~\ref{lem:path}}
We prove the statement simply by calculating the second derivative of the path-regularizer with respect to each parameter:
\begin{align*}
\kappa_{i}(\vec p)&=\frac{1}{2}\frac{\partial^2 \gamma^2_{\net}}{\partial p_i^2} =\frac{1}{2}\frac{\partial}{\partial p_i}\left(\frac{\partial }{\partial p_i}\sum_{\zeta \in \calP} \prod_{j=0}^{\length(\zeta)-1} w_{\zeta_{j}\rightarrow \zeta_{j+1}}^2\right)\\
&=\frac{1}{2}\frac{\partial}{\partial p_i}\left(\frac{\partial }{\partial p_i}\sum_{\zeta \in \calP} \prod_{j=0}^{\length(\zeta)-1} p_{\pi(\zeta_{j}\rightarrow \zeta_{j+1})}^2\right)
=\frac{1}{2}\sum_{\zeta \in \calP}\frac{\partial}{\partial p_i}\left(\frac{\partial }{\partial p_i} \prod_{j=0}^{\length(\zeta)-1} p_{\pi(\zeta_{j}\rightarrow \zeta_{j+1})}^2\right)\\
\end{align*}
Taking the second derivative then gives us both terms after a few calculations:
\begin{align*}
\kappa_{i}(\vec p)&=\frac{1}{2}\sum_{\zeta \in \calP}\frac{\partial}{\partial p_i}\left(\frac{\partial }{\partial p_i} \prod_{j=0}^{\length(\zeta)-1} p_{\pi(\zeta_{j}\rightarrow \zeta_{j+1})}^2\right)=\sum_{\zeta \in \calP}\frac{\partial}{\partial p_i}\left(p_i\sum_{e \in E_i} \vec 1_{e\in \zeta}\prod_{j=0 \atop e\neq (\zeta_j\rightarrow \zeta_{j+1}}^{\length(\zeta)-1}p_{\pi(\zeta_{j}\rightarrow \zeta_{j+1})}^2\right) \\
&= \sum_{\zeta \in \calP}\left[ p_i\frac{\partial}{\partial p_i}\left(\sum_{e \in E_i} \vec 1_{e\in \zeta}\prod_{j=0 \atop e\neq (\zeta_j\rightarrow \zeta_{j+1}}^{\length(\zeta)-1}p_{\pi(\zeta_{j}\rightarrow \zeta_{j+1})}^2\right) + \sum_{e \in E_i} \vec 1_{e\in \zeta}\prod_{j=0 \atop e\neq (\zeta_j\rightarrow \zeta_{j+1}}^{\length(\zeta)-1}p_{\pi(\zeta_{j}\rightarrow \zeta_{j+1})}^2 \right]\\
&=p_i^2\sum_{e1,e2\in E_i \atop e_1\neq e_2}\left[\sum_{\zeta \in \calP} \vec 1_{e_1,e_2\in \zeta} \prod_{j=0 \atop { e_1\neq (\zeta_j\rightarrow \zeta_{j+1}) \atop e_2\neq (\zeta_j\rightarrow \zeta_{j+1}) }}^{\length(\zeta)-1} p^2_{\pi(\zeta_j\rightarrow \zeta_{j+1})}\right]
+\sum_{e\in E_i}\left[\sum_{\zeta \in \calP} \vec 1_{e\in \zeta} \prod_{j=0 \atop e\neq (\zeta_j\rightarrow \zeta_{j+1})}^{\length(\zeta)-1} p^2_{\pi(\zeta_j\rightarrow \zeta_{j+1})}\right]\\
\end{align*}\qedwhite

\subsection{Proof of Theorem~\ref{thm:pathsgd-inv}}
Node-wise rescaling invariances for a feedforward network can be written as $\calT_\beta(\vec w)_{u\rightarrow v}=(\beta_v/\beta_u)w_{u\rightarrow v}$ for some $\beta$ where $\beta_v>0$ for internal nodes and $\beta_v=1$ for any input/output nodes. Any feasible invariance for a network with shared weights can also be written in the same form. The only difference is that some of $\beta_v$s are now tied to each other in a way that shared weights have the same value after transformation. First, note that since the network is invariant to the transformation, the following statement holds by an induction similar to Theorem~\ref{thm:net-inv} but in the backward direction:
\begin{equation}
\frac{\partial L}{\partial h_v}(\calT_\beta(\vec p)) = \frac{1}{\beta_v}\frac{\partial L}{\partial h_u}(\vec p) 
\end{equation}
for any $(u\rightarrow v)\in E$. Furthermore, by the proof of the Theorem~\ref{thm:net-inv} we have that for any $(u\rightarrow v)\in E$, $h_u(\calT_\beta(\vec p ) ) =  \beta_u h_u(\vec p)$. Therefore,
\begin{equation}
\frac{\partial L}{\partial \calT_\beta(\vec p)_i}(\calT_\beta(\vec p)) = \sum_{(u\rightarrow v)\in E_i}\frac{\partial L}{\partial h_v}(\calT_\beta(\vec p)) h_u(\calT_\beta(\vec p ) ) =  \frac{\beta_{u'}}{\beta_{v'}}\frac{\partial L}{\partial p_i}(\vec p) 
\end{equation}
where $(u'\rightarrow v')\in E_i$. In order to prove the theorem statement, it is enough to show that for any edge $(u\rightarrow v) \in E_i$, $\kappa_{i}(\calT_\beta(\vec p)) = (\beta_u/\beta_v)^2\kappa_{i}(\vec p)$ because this property gives us the following update:
\begin{equation*}
\calT_\beta(\vec p)_i - \frac{\eta}{\kappa_i(\calT_\beta(\vec p))} \frac{\partial L(\calT_\beta(\vec p))}{\partial \calT_\beta(\vec p)_i } = \frac{\beta_v}{\beta_u}p_i-
\frac{\eta}{(\beta_u/\beta_v)^2\kappa_{i}(\vec p)}\frac{\beta_{u}}{\beta_{v}}\frac{\partial L}{\partial p_i}(\vec p) = \calT_\beta(\vec p^+)_i
\end{equation*}
Therefore, it is remained to show that for any edge $(u\rightarrow v) \in E_i$ $v$, $\kappa_{i}(\calT_\beta(\vec p)) = (\beta_u/\beta_v)^2\kappa_{i}(\vec p)$. We show that this is indeed true for both terms $\kappa^{(1)}$ and $\kappa^{(2)}$ separately. 

We first prove the statement for $\kappa^{(1)}$. Consider each path $\zeta\in \calP$. By an inductive argument along the path, it is easy to see that multiplying squared weights along this path is invariant to the transformation:
\begin{equation*}
\prod_{j=0}^{\length(\zeta)-1} \calT_\beta(\vec p)^2_{\pi(\zeta_j\rightarrow \zeta_{j+1})} = \prod_{j=0}^{\length(\zeta)-1} p^2_{\pi(\zeta_j\rightarrow \zeta_{j+1})}
\end{equation*}
Therefore, we have that for any edge $e\in E$ and any $\zeta\in \calP$,
\begin{equation*}
\prod_{j=0 \atop e\neq (\zeta_j\rightarrow \zeta_{j+1})}^{\length(\zeta)-1} \calT_\beta(\vec p)^2_{\pi(\zeta_j\rightarrow \zeta_{j+1})}
= \left(\frac{\beta_u}{\beta_v}\right)^2 \prod_{j=0  \atop e\neq (\zeta_j\rightarrow \zeta_{j+1})}^{\length(\zeta)-1} p^2_{\pi(\zeta_j\rightarrow \zeta_{j+1})}
\end{equation*}
Taking sum over all paths $\zeta\in \calP$ and all edges $e=(u\rightarrow v) \in E$ completes the proof for $\kappa^{(1)}$. Similarly for $\kappa^{(2)}$, considering any two edges $e_1\neq e_2$ and any path $\zeta_\calP$, we have that:
\begin{equation*}
\calT_\beta(\vec p)_i^2\prod_{j=0 \atop { e_1\neq (\zeta_j\rightarrow \zeta_{j+1}) \atop e_2\neq (\zeta_j\rightarrow \zeta_{j+1}) }}^{\length(\zeta)-1} \calT_\beta(\vec p)^2_{\pi(\zeta_j\rightarrow \zeta_{j+1})} =  \left(\frac{\beta_v}{\beta_u}\right)^2 p_i^2 \left(\frac{\beta_u}{\beta_v}\right)^4\prod_{j=0 \atop { e_1\neq (\zeta_j\rightarrow \zeta_{j+1}) \atop e_2\neq (\zeta_j\rightarrow \zeta_{j+1}) }}^{\length(\zeta)-1} p^2_{\pi(\zeta_j\rightarrow \zeta_{j+1})}
\end{equation*}
where $(u\rightarrow v)\in E_i$. Again, taking sum over all paths $\zeta$ and all edges $e_1\neq e_2$ proves the statement for $\kappa^{(2)}$ and consequently for $\kappa^{(1)}+\kappa^{(2)}$.

\qedwhite
\subsection{Proof of Theorem~\ref{thm:pathsgd-cal}}
First, note that based on the definitions in the theorem statement, for any node $v$, $h_v(\tilde{\vec p})=\gamma^2_v(p)$ and therefore $g(\vec 1)=\gamma_\net^2(p)$. Using Lemma~\ref{lem:path}, main observation here is that for each edge $e\in E_i$ and each path $\zeta\in \calP$, the corresponding term in $\kappa^{(1)}$ is nothing but product of the squared weights along the path except the weights that correspond to the edge $e$:
$$
\vec 1_{e\in \zeta} \prod_{j=0 \atop e\neq (\zeta_j\rightarrow \zeta_{j+1})}^{\length(\zeta)-1} p^2_{\pi(\zeta_j\rightarrow \zeta_{j+1})}
$$
This path can therefore be decomposed into a path from input to edge $e$ and a path from edge $e$ to the output. Therefore, for any edge $e$, we can factor out the number corresponding to the paths that go through $e$ and rewrite $\kappa^{(1)}$ as follows:
\begin{equation}
\kappa^{(1)}(p)=\sum_{(u\rightarrow v)\in E_i} \left[\left(\sum_{\zeta\in \calP_{\In \rightarrow u} } \prod_{j=0}^{\length(\zeta)-1} p_{\pi(\zeta_{j}\rightarrow \zeta_{j+1})}^2\right)\left(\sum_{\zeta\in \calP_{v\rightarrow \Out} } \prod_{j=0}^{\length(\zeta)-1}p_{\pi(\zeta_{j}\rightarrow \zeta_{j+1})}^2 \right)\right]
\end{equation}
where $\calP_{\In\rightarrow u}$ is the set of paths from input nodes to node $v$ and $\calP_{v\rightarrow \Out}$ is defined similarly for the output nodes.

By induction on layers of $\calN(G,\pi,\tilde{\vec p})$, we get the following:
\begin{align}
\sum_{\zeta\in \calP_{\In \rightarrow u} } \prod_{j=0}^{\length(\zeta)-1} p_{\pi(\zeta_{j}\rightarrow \zeta_{j+1})}^2 &= h_u(\tilde{\vec p})\\
\sum_{\zeta\in \calP_{v\rightarrow \Out} } \prod_{j=0}^{\length(\zeta)-1}p_{\pi(\zeta_{j}\rightarrow \zeta_{j+1})}^2 &= \frac{\partial g(1)}{\partial h_v(\tilde{\vec p})}
\end{align}
Therefore, $\kappa^{(1)}$ can be written as:
\begin{equation}
\kappa^{(1)}(p)= \sum_{(u\rightarrow v)\in E_i} \frac{\partial g(1)}{\partial h_v(\tilde{\vec p})}h_u(\tilde{\vec p}) 
= \sum_{(u\rightarrow v)\in E_i} \frac{\partial g(1)}{\partial w'_{u\rightarrow v}} =  \frac{\partial g(1)}{\partial \tilde{p}_{i}}
\end{equation}
Next, we show how to calculate the second term, i.e. $\kappa^{(2)}$. Each term in $\kappa^{(2)}$  corresponds to a path that goes through two edges. We can decompose such paths and rewrite $\kappa^{(2)}$ similar to the first term:
\begin{align*}
\kappa^{(2)}(p)&=p_i^2\sum_{(u\rightarrow v)\in E_i \atop{ (u'\rightarrow v')\in E_i \atop (u\rightarrow v) \neq (u'\rightarrow v')} } \left[\left(\sum_{\zeta\in \calP_{\In \rightarrow u}}\prod_{j=0}^{\length(\zeta)} p_{\pi(\zeta_{j}\rightarrow \zeta_{j+1})}^2\right)\right.\\
&\left.\left(\sum_{\zeta\in \calP_{v \rightarrow u'}}\prod_{j=0}^{\length(\zeta)-1}p_{\pi(\zeta_{j}\rightarrow \zeta_{j+1})}^2 \right)\left(\sum_{\zeta\in \calP_{v'\rightarrow \Out}}\prod_{j=0}^{\length(\zeta)-1}p_{\pi(\zeta_{j}\rightarrow \zeta_{j+1})}^2\right)\right]\\
&= \sum_{(u\rightarrow v)\in E_i \atop{ (u'\rightarrow v')\in E_i \atop (u\rightarrow v) \neq (u'\rightarrow v')} } \tilde{p}_i\frac{\partial g(\vec 1)}{\partial h_{v'}(\tilde{\vec p})}\frac{\partial h_{u'}(\tilde{\vec p})}{\partial h_v(\tilde{\vec p})} h_u(\tilde{\vec p})
\end{align*}
where $\calP_{u\rightarrow v}$ is the set of all directed paths from node $u$ to node $v$.

\qedwhite

\end{document}